%% file: main.tex
\documentclass[10pt,twocolumn,letterpaper]{article}

\usepackage[pagenumbers]{cvpr} 

\input{preamble}
\input{math_commands}

\definecolor{cvprblue}{rgb}{0.21,0.49,0.74}
\usepackage[pagebackref,breaklinks,colorlinks,allcolors=cvprblue]{hyperref}

\def\paperID{xxxx}
\def\confName{CVPR}
\def\confYear{2026}

\title{ASAP-Textured Gaussians: Enhancing Textured Gaussians with Adaptive Sampling and Anisotropic Parameterization}

\author{
    Meng Wei$^{1}$ \quad
    Cheng Zhang$^{1}$ \quad
    Jianmin Zheng$^{2}$ \quad \\
    Hamid Rezatofighi$^{1}$ \quad
    Jianfei Cai$^{1}$
    \\
    $^{1}$Monash University \quad
    $^{2}$Nanyang Technological University
}

\begin{document}
\maketitle

\input{sec/0_abstract}
\input{sec/1_intro}
\input{sec/2_related}
\input{sec/3_preliminary}
\input{sec/3_method}
\input{sec/4_experiment}
\input{sec/5_conclusion}
{
    \small
    \bibliographystyle{ieeenat_fullname}
    \bibliography{main, foundation}
}

\input{sec/X_suppl.tex}

\end{document}

%% file: preamble.tex
\usepackage{colortbl}

\usepackage{array}
\usepackage{arydshln}
\definecolor{best}{HTML}{F39A9A}   
\definecolor{second}{HTML}{FDD5B1} 
\definecolor{third}{HTML}{FFF2CC}  


%% file: math_commands.tex
\usepackage{amsmath,amsfonts,bm}

\def\eqref#1{equation~\ref{#1}}

\def\1{\bm{1}}

\def\vmu{{\bm{\mu}}}

\def\vc{{\bm{c}}}

\def\vf{{\bm{f}}}
\def\vg{{\bm{g}}}

\def\vn{{\bm{n}}}

\def\vp{{\bm{p}}}

\def\vr{{\bm{r}}}
\def\vs{{\bm{s}}}
\def\vt{{\bm{t}}}
\def\vu{{\bm{u}}}

\def\mH{{\bm{H}}}

\def\mM{{\bm{M}}}

\def\mW{{\bm{W}}}

\DeclareMathAlphabet{\mathsfit}{\encodingdefault}{\sfdefault}{m}{sl}
\SetMathAlphabet{\mathsfit}{bold}{\encodingdefault}{\sfdefault}{bx}{n}

\def\gG{{\mathcal{G}}}

\def\gK{{\mathcal{K}}}



%% file: sec/0_abstract.tex
\begin{abstract}
Recent advances have equipped 3D Gaussian Splatting with texture parameterizations to capture spatially varying attributes, improving the performance of both appearance modeling and downstream tasks.
However, the added texture parameters introduce significant memory efficiency challenges.
Rather than proposing new texture formulations, we take a step back to examine the characteristics of existing textured Gaussian methods and identify two key limitations in common: 
(1) Textures are typically defined in canonical space, leading to inefficient sampling that wastes textures' capacity on low-contribution regions; and (2) texture parameterization is uniformly assigned across all Gaussians, regardless of their visual complexity, resulting in over-parameterization.
In this work, we address these issues through two simple yet effective strategies: adaptive sampling based on the Gaussian density distribution and error-driven anisotropic parameterization that allocates texture resources according to rendering error. Our proposed \textit{ASAP-Textured Gaussians}, short for Adaptive Sampling and Anisotropic Parameterization, significantly improve the quality–efficiency trade-off, achieving high-fidelity rendering with far fewer texture parameters.
\end{abstract}

%% file: sec/1_intro.tex
\section{Introduction}
\label{sec:intro}
3D Gaussian Splatting (3DGS)~\cite{3DGS_2023} has emerged as an efficient and effective representation for 3D scenes, offering real-time performance with photorealistic rendering quality. Originally developed for novel view synthesis~\cite{Yu2023MipSplatting} and 3D reconstruction~\cite{huang20242d}, it has since been extended to various downstream tasks by associating each Gaussian with task-specific attributes. Such extensions have enabled 3DGS to function as a general-purpose 3D backbone across applications, including dynamic scene modeling~\cite{wu20244DGS}, physical simulation~\cite{xie2023physgaussian}, and 3D scene understanding~\cite{li2025scenesplat}.

Recent works have explored enriching the expressive capacity of Gaussians by embedding spatially varying attributes through local texture parameterizations. These textured representations have been applied to encode appearance~\cite{xu2024texture, xu2024SuperGaussians,rong2025gstex}, geometry~\cite{svitov2025billboard,chao2025texturedgaussians}, and materials~\cite{younes2025textureplat_ref}, significantly broadening the representational scope of 3DGS. In particular, textured Gaussians~\cite{chao2025texturedgaussians,svitov2025billboard,younes2025textureplat_ref} for appearance have proven effective in capturing high-frequency details and structural variations through RGBA textures, surpassing the limitations of uniform color and ellipsoidal shape.

However, while textured Gaussians improve image fidelity, they also introduce new memory efficiency challenges: Attaching a dedicated texture to each Gaussian introduces significant memory overhead and complicates the balance of rendering quality and resource usage. This raises a central question: \textit{Can we improve the utilization of texture resources in textured Gaussian Splatting without sacrificing rendering quality?} Addressing this question is critical for advancing the scalability and generalizability of textured Gaussians.

In this work, rather than proposing new texture formulations, we take a step back to analyze the properties of existing textured Gaussian representations and identify two fundamental limitations in common. 1) First, most methods treat the Gaussian's canonical space as its texture space, which leads to inefficient sampling—many texture samples are assigned to regions with negligible visual contribution. 2) Second, texture parameterization is typically fixed and uniformly distributed across all Gaussians, regardless of each primitive's visual complexity or importance. This uniformity can lead to over-parameterization in simple areas, while under-representing regions that require more detail, ultimately reducing efficiency.

To address these limitations, we introduce two simple yet effective techniques: adaptive sampling and error-driven anisotropic parameterization. Adaptive sampling warps the texture coordinate space according to the Gaussian's density distribution, concentrating samples where the rendered contribution is higher. Error-driven anisotropic parameterization allocates resolution to each texture based on its estimated rendering error, allowing both resolution and aspect ratio to adapt to content complexity. We refer to our representation as \textbf{ASAP-Textured Gaussians}, short for \textit{Adaptive Sampling and Anisotropic Parameterization}. By integrating these two techniques, our method preserves the expressive power of textured Gaussians while significantly improving parameter efficiency. ASAP-Textured Gaussians support flexible quality–efficiency trade-offs, achieving comparable or even superior rendering quality with far fewer texture parameters than prior approaches.

Our contributions can be summarized as follows:
\begin{itemize}
    \item We propose an adaptive sampling strategy that warps texture coordinates based on the Gaussian density, improving sampling efficiency for texture utilization.
    \item We introduce an error-driven anisotropic parameterization method that allocates texture resolution and aspect ratio based on rendering error, offering efficient texture resource allocation.
    \item Experiments on multiple datasets demonstrate that ASAP-Textured Gaussians achieve comparable or even superior rendering quality with significantly fewer texture parameters compared to existing methods.
\end{itemize}

%% file: sec/2_related.tex
\section{Related Work}\label{sec:related}
\paragraph{Neural Radiance Fields.}
Neural radiance fields (NeRF)~\cite{mildenhall2021nerf} have established a differentiable volume rendering paradigm for photorealistic novel view synthesis, inspiring various advancements in efficiency~\cite{yu2021plenoctrees, 3DGS_2023,muller2022instant}, quality~\cite{barron2021mip,barron2023zipnerf}, and generalizability~\cite{chen2024mvsplat,yu2021pixelnerf}.
Among these, 3D Gaussian Splatting (3DGS)~\cite{3DGS_2023} has achieved high-fidelity synthesis at interactive rates through an efficient splat-based rasterization.
It models 3D scenes as a collection of 3D Gaussians and gradually optimizes their geometric and appearance parameters via differentiable rendering.
Built upon 3DGS, subsequent works further enhance robustness and scalability through stable training~\cite{kheradmand20243d}, anti-aliasing~\cite{Yu2023MipSplatting}, and geometry-aware reconstruction~\cite{huang20242d,chen2024pgsr}.
Recent works have also leveraged 3DGS as a versatile 3D backbone and augmented per-Gaussian attributes for semantics~\cite{li2025scenesplat,ye2023gaussian, xie2025unigs}, dynamics~\cite{wu20244DGS}, and physics~\cite{xie2023physgaussian,shi2023gir}, extending the vanilla Gaussian set to general-purpose 3D scene representation.

\paragraph{Textured Gaussian Splatting.}
To enrich the per-primitive expressiveness beyond spatially uniform attributes, recent works augment Gaussian primitives with local texture parameterizations. Early attempts, such as MLP-based texture mappings~\cite{xu2024texture}, demonstrate the feasibility of spatially varying appearances but struggle to scale to complex, scene-level data. Subsequent approaches build upon the 2D Gaussian Splatting (2DGS) framework~\cite{huang20242d}, which defines a local tangent plane for each Gaussian, allowing textures to be attached and sampled via UV coordinates aligned to the Gaussian’s principal axes. Examples include GSTex~\cite{rong2025gstex}, which adopts learned UV mappings, and Gaussian Billboards~\cite{weiss2024gaussianbillboards}, which employ fixed square textures for simplicity. SuperGaussians~\cite{xu2024SuperGaussians} further explore alternative kernel functions for texture representation, improving fidelity at the cost of higher complexity. TextureSplat~\cite{younes2025textureplat_ref} extends textures to model specular effects and employs texture atlases for acceleration. Recent attempts to generalize the geometry beyond ellipsoids—such as BBSplat~\cite{svitov2025billboard} and Textured Gaussians~\cite{chao2025texturedgaussians}—achieve strong visual quality but require substantially larger memory footprints.
Although these methods demonstrate the potential of textured representations, how to solve the memory efficiency challenge is an open-ended problem. Our work addresses this limitation through adaptive sampling and anisotropic texture parameterization, improving texture efficiency without compromising quality. Our design is orthogonal to compression-based methods such as vector quantization or ZIP-based packing~\cite{svitov2025billboard}, and can be combined with them for further memory savings.

%% file: sec/3_preliminary.tex
\section{Preliminaries} \label{sec:preliminary}

\begin{figure*}[t]
  \centering
  \includegraphics[width=0.95\linewidth]{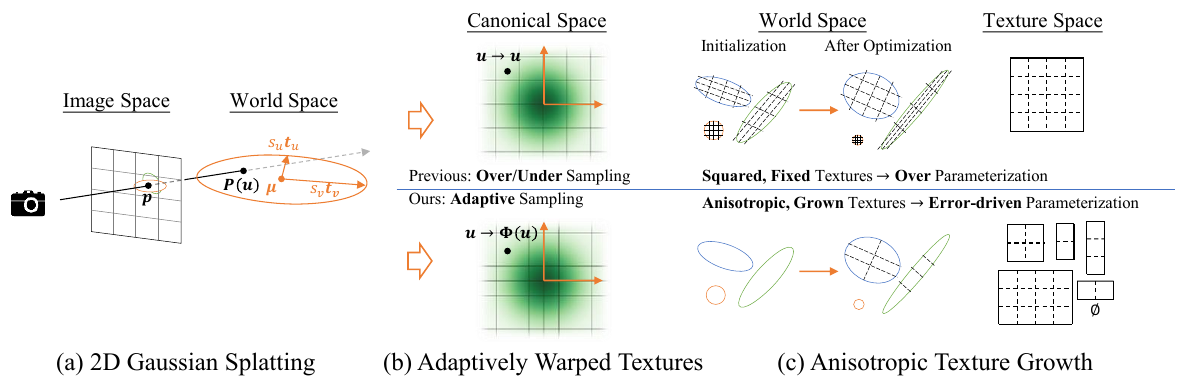}
  \caption{\textbf{Method Overview.}
  We introduce two key techniques to enhance the texture representation for (a) 2D Gaussian Splatting. (b) Adaptively Warped Textures: We introduce warping operations to align textures with the 2D Gaussian distribution, enabling more efficient usage of texture resources. Grid lines here correspond to the texture space coordinates visualized in the canonical space. (c) Anisotropic Texture Growth: We propose an error-driven texture resolution growth strategy that dynamically adjusts the texture resolution based on the learning status, allowing details to be captured where necessary while avoiding over-allocation of resources.}
  \label{fig:method}
\end{figure*}

\paragraph{2D Gaussian Splatting (2DGS).} 2DGS~\cite{huang20242d} represents scenes using flattened 2D Gaussians, enabling accurate surface reconstruction and providing a natural foundation for texture definition.
Each 2D Gaussian is parameterized as $\{ \vmu, \vr, \vs, o, \vc \}$, where $\vmu \in \mathbb{R}^3$ denotes the spatial mean, $\vr \in \mathbb{R}^{4}$ the rotation quaternion, $\vs \in \mathbb{R}^{2}$ the scaling factor, $o$ the opacity, and $\vc$ the color.
The 2D Gaussian is defined as a local tangent plane in 3D space, spanned by the first two orthogonal basis vectors $\{\vt_u, \vt_v\}$ from the corresponding rotation matrix, while the last basis vector represents the plane's normal vector $\vn$.

To render an image, the 2D Gaussians are first rasterized onto the image plane and sorted based on their depth values. Then, for each image coordinate $\vp \in \mathbb{R}^2$, the corresponding ray-splat intersection $\vu(\vp)$ is computed in the local canonical space of the 2D Gaussian.
Finally, the pixel color is obtained by alpha blending all 2D Gaussians covering the current pixel, from front to back: 
\begin{equation}
  C(\vp) = \sum_{i=1} \vc_i o_i \gG_i(\vu(\vp)) \prod_{j=1}^{i-1} \left(1 - o_j \gG_j(\vu(\vp))\right),
\end{equation}
where $i$ denotes the 2D Gaussian index in the depth-sorted order, and $\gG$ the standard 2D Gaussian function.

\paragraph{Mappings in 2DGS.}
Considering the 2D Gaussian with the geometric parameters $\{ \vmu, \vt_u, \vt_v, \vs \}$, where $\{\vt_u, \vt_v\}$ denotes the orthogonal principal axes, and $\vs = (s_u, s_v)$ the corresponding scales, any world-space points $\mathbf{P}$ on the local plane of the 2D Gaussian can be expressed as:
\begin{equation}
    \mathbf{P}(\vu) = \vmu + s_u \vt_u u + s_v \vt_v v,
\end{equation}
where $\vu = (u, v)$ denotes the local canonical coordinates on the 2D Gaussian's tangent plane.
Compactly, the mapping from the local canonical space to the world space could be represented as:
\begin{align}
    &\mathbf{P}(\vu) = \mH (u, v, 1, 1)^\top, \\
    &\text{where } \mH = \begin{bmatrix}
    s_u \vt_u & s_v \vt_v & \mathbf{0} & \vmu \\
    0 & 0 & 0 & 1
    \end{bmatrix} \in \mathbb{R}^{4\times4}
\end{align}
is the homogeneous transformation matrix from the canonical space to the world space.

The mapping between the camera space and the local canonical space of the 2D Gaussian is thus:
\begin{equation}
  \vr = (r_x z, r_y z, z, z)^\top = \mW\mH(u, v, 1, 1)^\top,
\end{equation}
where $\vr$ denotes a homogeneous ray exmitting from the pixel $(r_x, r_y)$ and intersecting with the splat at the depth $z$, $\mW \in \mathbb{R}^{4\times4}$ is the world-to-camera transformation matrix.
Inversely, $\vu = (\mW \mH)^{-1} \vr$, where $\mM = (\mW \mH)^{-1}$ is usually pre-computed for acceleration and named as the ``ray-transform'' matrix.
For more details, please refer to~\cite{huang20242d}.

%% file: sec/3_method.tex
\section{Method}
Our \textbf{ASAP-Textured Gaussian} builds upon the 2DGS framework and adopts the widely used texture-map representation for its simplicity and flexibility.
We improve the efficiency and scalability of textured 2DGS by addressing two fundamental limitations: redundant sampling within each primitive and over-parameterization across the scene.
To this end, we introduce two complementary techniques: (1) \emph{adaptive sampling} (Sec.~\ref{sec:warp}), which aligns sampling density with each Gaussian's mass distribution for intra-primitive efficiency, and (2) \emph{anisotropic texture resolution growth} (Sec.~\ref{sec:tex_growth}), which progressively adjusts texture size and aspect ratio based on gradient statistics for scalability.
An overview of our method is shown in Fig.~\ref{fig:method}.

\subsection{Adaptive Sampling for Textured 2D Gaussians} \label{sec:warp}
As discussed in Sec.~\ref{sec:preliminary}, the 2DGS framework defines the ray–splat intersections $\vu(\vp) = (u, v)$ in the canonical space of each Gaussian. This formulation naturally provides a convenient \emph{uv}-parameterization for associating textures with individual primitives.
Existing textured-Gaussian methods adopt this parameterization by either attaching explicit texture maps $\vc(u, v)$~\cite{chao2025texturedgaussians,svitov2025billboard} or learning MLPs $\gK(u, v)$~\cite{xu2024SuperGaussians,xu2024texture} to model spatially-varying appearances. While straightforward, these approaches implicitly assume that the \textbf{canonical space} is equivalent to the \textbf{texture space}.

However, this assumption overlooks the differences between spaces, caused by the non-uniform opacity distribution of the 2D Gaussian. As illustrated in Fig.~\ref{fig:method}~(b), directly using the canonical space as the texture space produces uniformly distributed texture samples, whereas their contribution to rendering decays exponentially from the center. Consequently, regions near the Gaussian tails—where opacity and color weights are negligible—still consume a large fraction of textures, while areas near the center—which contribute significantly to rendering—desire higher sampling densities to capture fine details. Such an imbalance results in inefficient usage of textures, motivating the need for warping functions that better align texture sampling densities with the 2D Gaussian's support.

\paragraph{Warping to align with the 2D Gaussian distribution.}
To address the inefficient usage of textures, we introduce warping functions that map the canonical coordinates $\vu = (u, v)$ to a mass-aware texture domain $\tilde{\vu} = (\tilde{u}, \tilde{v})$, such that the texture sampling density follows the Gaussian's local probability density.
Formally, we want to define the warping function, $\Phi(\vu) = \tilde{\vu}$, according to the cumulative distribution function (CDF) of the 2D Gaussian.

We consider two practical variants of warping functions:
(1) \emph{Axis-wise CDF warping}, which separately warps each axis, based on the 1D Gaussian CDF; and
(2) \emph{Radial CDF warping}, which warps the radial distance from the Gaussian center, respecting the radially symmetric nature of the canonical Gaussian.

The axis-wise CDF warping $\Phi_{\text{axis}}$ is defined as:
\begin{align}\label{eq:axis}
  \Phi_{\text{axis}}(a) = \frac{1}{2} \left( 1 + \mathrm{erf}\left(\frac{a}{\sqrt{2}}\right) \right),
\end{align}
where $\mathrm{erf}(\cdot)$ is the error function of the 1-D Gaussian distribution, measuring the cumulative probability mass.

The radial CDF warping $\Phi_{\text{radial}}$ is defined as:
\begin{align}
  r &= \sqrt{u^2 + v^2}, \\
  \tilde{r} &= 1 - e^{-\frac{r^2}{2}}, \\ \label{eq:radial}
  \Phi_{\text{radial}}(a) &= \frac{\tilde{r}}{r} a,
\end{align}
where $r$ specifies the radial distance of $\vu$ from the center and $\tilde{r}$ is the warped radial distance using the Rayleigh CDF.

Both warping functions redistribute texel sampling densities according to the Gaussian's mass distribution (either marginally or radially), better matching the opacity fall-off.
The final color contribution is thus $\vc(\tilde{\vu})$, avoiding wasted texture capacity in low-contribution regions and enabling detail representation around the center.

\paragraph{Discussions.}
The proposed warping serves as a bridge between the canonical and texture spaces and is agnostic to the appearance representation $\vc(\tilde{\vu})$, supporting both explicit texture maps and learned kernels or MLPs.
While the warping is well-defined for RGB textures, it remains practical for RGBA settings as well.
Through interpreting the Gaussian distribution as a prior on opacity, the warping functions thus effectively perform importance-based sampling: the texel density is concentrated in regions of higher expected contribution, guided by the Gaussian mass.
Overall, the warping function provides a principled mechanism to allocate texture capacity within each Gaussian, thereby improving texture utilization regardless of underlying appearance models.

\subsection{Anisotropic Texture Resolution Growth} \label{sec:tex_growth}
While warping improves sampling efficiency within each Gaussian, existing textured-Gaussian methods still suffer from fixed per-Gaussian texture parameterizations across the entire scene. As illustrated in Fig.~\ref{fig:method} (c), such uniform parameterizations disregard the spatially varying complexity of scene appearance: regions with smooth color variations or small geometric support receive the same texel budget as highly detailed or large surfaces, resulting in redundant memory and unnecessary computation.

To address this limitation, we propose an \emph{error-driven anisotropic texture resolution growth} strategy that progressively allocates texture capacity across Gaussians during training.
Inspired by the adaptive density control in~\cite{3DGS_2023}, we monitor the magnitude of texture gradients within each Gaussian as an indicator of representational adequacy, and adaptively adjust its texture size and aspect ratio to match local appearance complexity.
We follow~\cite{3DGS_2023,chao2025texturedgaussians} and use the photometric loss $\mathcal{L}$ between rendered and ground-truth images as the supervision signal, defined as:
\begin{equation}
  \mathcal{L} = (1 - \lambda_{\text{SSIM}}) \mathcal{L}_{1} + \lambda_{\text{SSIM}}\mathcal{L}_{\text{SSIM}}.
\end{equation}

\paragraph{Adaptive axis growth.}
To enable anisotropic resolution adaptation, we analyze the gradient statistics of textures along each axis.
Intuitively, large gradients along one axis indicate insufficient texture resolution in that direction, motivating additional texels to be allocated.
Formally, for each Gaussian with the texture size $(T_u, T_v)$, we accumulate the row- and column-wise gradients of texel values (either colors or features) with respect to the rendering loss $\mathcal{L}$:
\begin{align}
   \vg_u[i] &= \sum_{j=0}^{T_v-1} \left|\frac{\partial \mathcal L}{\partial \vf(i,j)}\right|,\qquad
   i=0,\dots,T_u-1, \\
   \vg_v[j] &= \sum_{i=0}^{T_u-1} \left|\frac{\partial \mathcal L}{\partial \vf(i,j)}\right|,\qquad
   j=0,\dots,T_v-1,
\end{align}
where $\vf(i,j)$ denotes the texel value at coordinate $(i,j)$.
The aggregated vectors  $\vg_u \in \mathbb{R}^{T_u}$ and $\vg_v \in \mathbb{R}^{T_v}$ quantify the per-axis update pressure, serving as indicators for directional texture growth in subsequent steps.

To ensure robustness, the per-axis gradients are accumulated across multiple training iterations and views.
For each Gaussian $g$, we record its visibility count $n_g$, accounting for observation frequencies.
After accumulation, the directional gradient pressures are averaged over both the visibility count and the per-axis texel size:
\begin{align}
   \bar \vg_u &= \frac{1}{n_g T_v}\sum_{i=0}^{T_u-1} \vg_u[i], &
   \bar \vg_v &= \frac{1}{n_g T_u}\sum_{j=0}^{T_v-1} \vg_v[j].
\end{align}
This averaging mitigates bias toward axes that currently contain more texels or Gaussians that appear in more views.

\paragraph{Implementation details of growth decision.}
A Gaussian expands when the corresponding averaged gradient magnitude $\bar \vg_u$ or $\bar \vg_v$ exceeds a predefined gradient threshold $\tau_\text{tex}$.
Our growth checks are performed periodically—every $N_\text{tex}$ iterations—to suppress transient gradient fluctuations, and the maximum number of growth steps is limited by $N_\text{max}$ to control memory usage and allow stable fine-tuning thereafter.
Whenever a texture expands, it is resampled to the new resolution using bilinear interpolation, ensuring smooth parameter transfer and continuity without introducing rendering artifacts.

\paragraph{Texture Initialization.}
To maximize the benefits of adaptive growth, we initialize Gaussians without textures and accumulate gradients over base appearance attributes to determine whether texture activation is necessary.
Once activated, the texture is initialized with a small anisotropic grid, $(1\times2)$ or $(2\times1)$, where the aspect ratio is determined by the Gaussian's relative axis scales.
Some concurrent work~\cite{anonymous2025atg} explores texture resolution adaptation based on geometric size and lacks a statistical measure of representational adequacy along each axis.
In contrast, our gradient-driven criterion is more precise and appearance-aware: it directly reflects the rendering error propagated through each texel.
For instance, Gaussians with nearly uniform colors may remain low-resolution or untextured under our approach, even when spatially elongated, while visually complex regions naturally trigger anisotropic texture expansion.

\paragraph{CUDA-Efficient Anisotropic Texture Pipeline}
A key advantage of 3DGS lies in its high-performance CUDA rasterization pipeline. To ensure that our anisotropic textures remain fully compatible with this parallel rendering framework, we design an implementation that preserves GPU-friendly data access patterns and avoids divergence across kernels. Specifically, we determine a global maximum texture resolution and employ a differentiable bilinear resampling scheme that maps each Gaussian’s stored anisotropic texture to a uniform texture grid passed to the CUDA kernels. This design maintains efficient parallelism and prevents irregularities caused by texture size variability. During inference, the resampling is performed once and does not alter the standard textured-Gaussian pipeline. As a result, our anisotropic parameterization introduces only a minimal computational overhead during training, as verified in our experiments. Moreover, complementary acceleration techniques such as texture ATLAS representations~\cite{younes2025textureplat_ref, purnomo2004seamless} can be integrated for additional speedups and compression.

In summary, our \textbf{ASAP-Textured Gaussians} enhance texture efficiency from both intra- and cross-primitive perspectives.
The proposed \emph{adaptive warping} aligns sampling density with each Gaussian's mass distribution to eliminate redundant texel usage, while the \emph{error-driven anisotropic texture growth} adaptively allocates texture capacity across Gaussians according to gradient-derived error statistics.
Together, these components form a unified framework that achieves high-fidelity rendering with substantially improved texture efficiency and scalability.

%% file: sec/4_experiment.tex
\section{Experiments}

\input{tab/main.tex}
\input{figure/main.tex}
\input{tab/ablate_warp.tex}
\input{tab/ablate_radial.tex}
\input{tab/effeciency.tex}

\subsection{Experimental Setup}

\paragraph{Datasets and Metrics.}
Following the common practice in novel view synthesis~\cite{barron2021mip,3DGS_2023}, we conducted experiments on three widely-used datasets, including the \emph{Mip-NeRF 360} dataset~\cite{barron2022mipnerf360} (7 scenes), the \emph{Tanks and Temples} dataset~\cite{Knapitsch2017} (2 scenes), and the \emph{Deep Blending} dataset~\cite{DeepBlending2018} (2 scenes). The data preprocessing and train/test splits follow the original settings in 3DGS~\cite{3DGS_2023}.
We evaluated the rendering quality using three standard metrics, including the Peak Signal-to-Noise Ratio (PSNR), the Structural Similarity Index Measure (SSIM)~\cite{Wang2004SSIM}, and the Learned Perceptual Image Patch Similarity (LPIPS)~\cite{zhang2018lpips}. We also report the number of Gaussians and the model size (in MB) to evaluate the efficiency of different methods.

\paragraph{Baselines.}
We compared our method with four closely related 3DGS-based neural rendering methods:
1) 2D Gaussian Splatting (\underline{2DGS})~\cite{huang20242d}, the backbone of many texture-based methods; 2) \underline{Textured Gaussians} (\underline{TexGau})~\cite{chao2025texturedgaussians}, the state-of-the-art approach for textured Gaussian representation that achieves a balanced trade-off between efficiency and quality; 3) \underline{BBSplat}~\cite{svitov2025billboard} and 4) \underline{SuperGaussian}~\cite{xu2024SuperGaussians} (\underline{SuperGau}), two recent methods that improve the expressiveness of textured Gaussians.
All methods were retrained using their publicly available implementations.
For 2DGS, we retrained the model without geometric regularization for improved rendering quality and applied the MCMC strategy~\cite{kheradmand20243d} to control the exact Gaussian count, consistent with TexGau and BBSplat.
We used 2DGS\textsuperscript{\textdagger} to denote this modified version.
For TexGau, its released implementation is built on the 2DGS framework rather than 3DGS as stated in the paper; we denote this version as TexGau\textsuperscript{\textdagger}.
To isolate the effect of textured Gaussians, we retrained BBSplat without its sky-box strategy from the supplementary material, denoted as BBSplat\textsuperscript{\textdagger}.

\paragraph{Implementation details.}
Following TexGau, we used identical initialization, training schedule, learning rates, and optimizers for a fair comparison.
Specifically, both our ASAP-Textured Gaussians and TexGau shared exactly the same initialization of 30K-step pre-trained 2DGS and were fine-tuned for another 30K steps with textures. All other baselines were also trained for 60K steps to ensure fairness. All methods used the same order of Spherical Harmonics for view-dependent color modeling.
The gradient thresholds for our texture growth were set to $\tau_\text{base}=4\times10^{-6}$ and $\tau_\text{tex}=2\times10^{-7}$. Gradients were accumulated over 100 steps before evaluating the growth condition, which was applied periodically until 15K steps.
Thanks to our efficient texture parameterization, we could use detailed texture resolutions up to $8\times8$ per Gaussian without notable memory overhead.
Unless otherwise stated, all experiments shared identical settings and were conducted on NVIDIA A100 GPUs.
Our code and trained models will be released to support future research.

\subsection{Overall Results}
We compare our ASAP-Textured Gaussians against baseline methods under the same number of Gaussians, ranging from 100K to 500K. Model size is reported as the total number of parameters before any quantization or ZIP-based compression~\cite{svitov2025billboard}. We use the base 2DGS model size as the reference and report the relative overhead in percentage. Unless stated otherwise, axis-wise CDF warping is used for its simplicity; a detailed comparison with radial CDF warping is provided in Sec.~\ref{sec:ablation}.

As shown in Tab.~\ref{tab:main}, our method consistently achieves comparable or better rendering quality while requiring fewer parameters, as reflected in the memory footprint (MB). BBSplat introduces more than an order of magnitude additional texture parameters (over 15×), so we report it only as a reference. SuperGau exhibits a similar model size to ours but performs significantly worse when the Gaussian count is small (100K and 200K). The quality–size trade-offs in Fig.~\ref{fig:main_plot} further highlight this trend: points in the upper-right corner indicate better trade-offs. Our ASAP-Textured Gaussians achieve a good trade-off across all datasets.

From Tab.~\ref{tab:main}, we observe that under a small Gaussian budget (e.g., 100K), the scene is under-represented, leading to higher modeling error. Consequently, our gradient-based, error-driven allocation assigns more texture resources, which results in a slightly higher proportion of added parameters in this low-budget regime. As the number of Gaussians increases, the overall rendering quality increases, and our adaptive mechanism becomes more selective, as reflected in the decreasing percentage of added parameters.
Fig.~\ref{fig:main_qualitative} shows qualitative comparisons on representative scenes from the three datasets. Our method achieves comparable results when using the same number of Gaussians, but with fewer parameters. When the model size is similar, our method could synthesize visually appealing results, indicating our methods maintain a good trade-off between image quality and memory efficiency. For more results, please refer to the Supplementary.

We also report the additional texture map training time of the baseline and our method on the MipNeRF-360 dataset. As Tab.~\ref{tab:time_training} illustrates, our implementation is GPU-friendly, introducing small training overhead.

\subsection{Ablation studies.}\label{sec:ablation}
\paragraph{Comparisons between axis-wise and radial CDF warping functions.}
Tab.~\ref{tab:ablation_radial_warp} shows that the approximate axis-wise CDF warping yields performance improvements comparable to the precise radial CDF warping.
As reported in Table~\ref{tab:time_rendering}, both warping functions introduce negligible overhead at inference time and preserve real-time rendering performance.
We adopt axis-wise CDF warping for the remaining experiments, considering its implementation simplicity.

\paragraph{Varied texture resolution and number of Gaussians.}
We further evaluate the effectiveness of the proposed warping functions under different texture resolutions and Gaussian counts on the Mip-NeRF 360 dataset.
All settings are kept identical across experiments except for the warping function, ensuring a fair comparison.
As shown in Tab.~\ref{tab:ablation_warp}, incorporating the axis-wise warping function consistently improves reconstruction quality across all texture resolutions and Gaussian numbers.
A notable observation is that, with warping, the performance of lower-resolution textures can match or even surpass that of higher-resolution textures without warping.
This highlights a key limitation of previous approaches: to achieve comparable fidelity near the high-contribution center of each Gaussian, they rely on uniformly allocating excessively large textures, which results in redundant sampling and unnecessary memory cost.
In contrast, warping concentrates sampling density where it matters most, enabling significantly more efficient use of texture capacity.

\paragraph{RGB vs RGBA textures.}
We also evaluate the impact of our warping operations on RGB textures across different numbers of Gaussians. As shown in Tab.~\ref{tab:ablation_rgb}, the warping operation consistently improves the rendering quality, regardless of the underlying textures, demonstrating its robustness and potential generalizability to other parameterizations.

\paragraph{Limitations and Future Work.}
While our method achieves strong performance, several practical aspects present opportunities for further improvement.
First, the anisotropic texture parameterization introduces a modest amount of training overhead, though this has no impact on inference speed. Our implementation also trades a small amount of runtime memory for GPU compatibility; incorporating texture ATLAS representations~\cite{younes2025textureplat_ref, purnomo2004seamless} could offer additional speedups and memory savings.
Second, although our error-driven texture growth strategy is effective, integrating Monte Carlo–based adaptive sampling~\cite{kheradmand20243d} presents a promising direction for further enhancing both efficiency and reconstruction quality.

%% file: tab/main.tex
\begin{figure*}[htbp]
    \centering
    \includegraphics[width=0.85\linewidth]{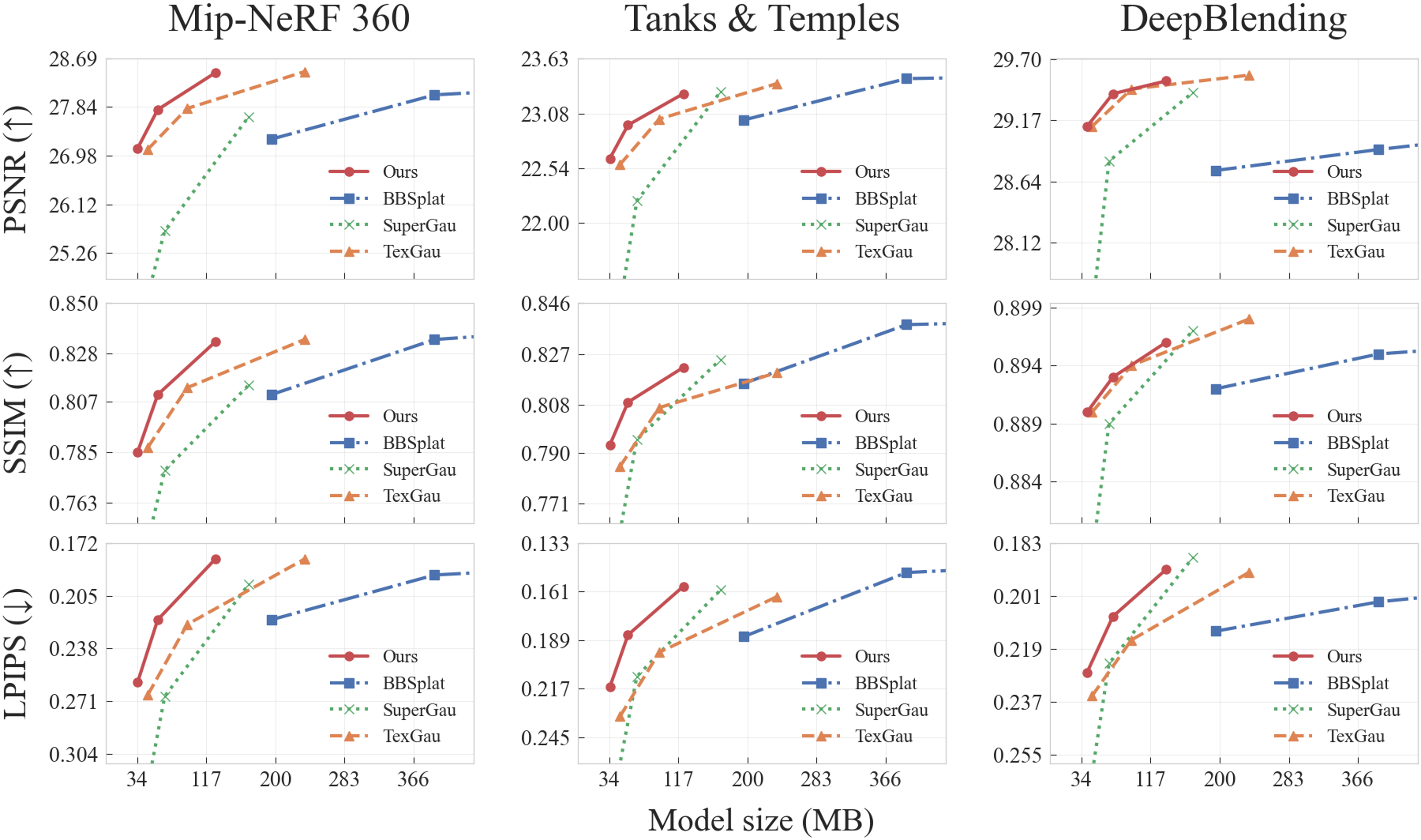} \\
    \caption{\textbf{Rendering Quality vs. Model Size.}
    Rendering quality (PSNR, SSIM, LPIPS) plotted against model size (MB) on three datasets.
    For clarity, we zoom in to exclude outliers with exceptionally large memory footprints or poor performance.
    Across all datasets, our ASAP-Textured Gaussians consistently achieve superior trade-offs between rendering quality and model size compared to prior approaches.}
    \label{fig:main_plot}
\end{figure*}

\begin{table*}[t] \footnotesize 
    \caption{\textbf{Quantitative comparisons.} Our method achieves comparable or even better rendering quality with fewer parameters than baseline textured Gaussians methods. BBsplat uses significantly larger memory and is included here only for reference. The model size here is measured as the parameter size of the entire model.}
    \label{tab:main}
    \centering
    \begin{tabular}{l l ccc ccc ccc}
        \toprule
        \# Gaussians & Mem & \multicolumn{3}{c}{ Mip-NeRF 360} &  \multicolumn{3}{c}{ Tanks \& Temples}&  \multicolumn{3}{c}{ DeepBlending} \\
        \cmidrule(rr){1-1} \cmidrule(lr){2-2} \cmidrule(lr){3-5}  \cmidrule(lr){6-8}  \cmidrule{9-11}
        Method
        & \scriptsize MB (\%)
        & \scriptsize PSNR $\uparrow$    & \scriptsize SSIM $\uparrow$  & \scriptsize LPIPS $\downarrow$
        & \scriptsize PSNR $\uparrow$    & \scriptsize SSIM $\uparrow$  & \scriptsize LPIPS $\downarrow$
        & \scriptsize PSNR $\uparrow$    & \scriptsize SSIM $\uparrow$  & \scriptsize LPIPS $\downarrow$
        \\
        \midrule
        \multicolumn{2}{l}{\# 100K} \\
        \midrule
            2DGS\textsuperscript{\textdagger}
            & 22.51 (100\%) 
            & 26.71 & 0.775 & 0.283
            & 22.22 & 0.779 & 0.241
            & 28.66 & 0.882 & 0.253 \\
            \rowcolor{gray!20}
            BBSplat\textsuperscript{\textdagger} (ref.)
            & 391 (+1638\%)
            & 28.05 & 0.834 & 0.192
            & 23.43 & 0.838 & 0.150
            & 28.92 & 0.895 & 0.203 \\
            SuperGau
            & 33.6 (+49\%)
            & 23.78 & 0.726 & 0.361
            & 20.51 & 0.738 & 0.293
            & 26.67 & 0.870 & 0.286 \\
            TexGau\textsuperscript{\textdagger}
            & 46.9 (+108\%)
            & \cellcolor{second}27.09 & \cellcolor{best}0.787 & \cellcolor{second}0.267
            & \cellcolor{second}22.58 & \cellcolor{second}0.785 & \cellcolor{second}0.233
            & \cellcolor{best}29.12 & \cellcolor{best}0.890 & \cellcolor{second}0.235 \\
            Ours
            & 34.6 (+54\%)
            & \cellcolor{best}27.11 & \cellcolor{second}0.785 & \cellcolor{best}0.259
            & \cellcolor{best}22.64 & \cellcolor{best}0.793 & \cellcolor{best}0.216
            & \cellcolor{best}29.12 & \cellcolor{best}0.890 & \cellcolor{best}0.227 \\
        \arrayrulecolor{black}
        \midrule
        \multicolumn{2}{l}{\# 200K} \\
        \midrule
            2DGS\textsuperscript{\textdagger}
            & 45.0 (100\%) 
            & 27.50 & 0.804 & 0.235
            & 22.67 & 0.800 & 0.202
            & 29.13 & 0.891 & 0.226 \\
            \rowcolor{gray!20}
            BBSplat\textsuperscript{\textdagger} (ref.)
            & 781 (+1636\%)
            & 28.37 & 0.844 & 0.181
            & 23.48 & 0.841 & 0.141
            & 29.24 & 0.897 & 0.192 \\
            SuperGau
            & 67.1 (+49\%) 
            & 25.65 & 0.777 & 0.268
            & 22.22 & 0.795 & 0.210
            & 28.82 & 0.889 & 0.224 \\
            TexGau\textsuperscript{\textdagger}
            & 93.6 (+108\%)
            & \cellcolor{best}27.81 & \cellcolor{best}0.813 & \cellcolor{second}0.223
            & \cellcolor{best}23.03 & \cellcolor{second}0.807 & \cellcolor{second}0.196
            & \cellcolor{best}29.44 & \cellcolor{best}0.894 & \cellcolor{second}0.216 \\
            Ours
            & 60.7 (+35\%)
            & \cellcolor{second}27.79 & \cellcolor{second}0.810 & \cellcolor{best}0.220
            & \cellcolor{second}22.97 & \cellcolor{best}0.809 & \cellcolor{best}0.186
            & \cellcolor{second}29.40 & \cellcolor{second}0.893 & \cellcolor{best}0.208 \\
        \midrule
        \multicolumn{2}{l}{\# 500K} \\
        \midrule
            2DGS\textsuperscript{\textdagger}
            & 113 (100\%) 
            & 28.18 & 0.828 & 0.190
            & 23.09 & 0.817 & 0.169
            & 29.22 & 0.896 & 0.202 \\
            SuperGau
            & 168 (+48\%)
            & 27.66 & 0.814 & 0.198 
            & \cellcolor{second}23.30 & \cellcolor{best}0.825 & \cellcolor{second}0.160
            & 29.41 & \cellcolor{second}0.897 & \cellcolor{best}0.188 \\
            TexGau\textsuperscript{\textdagger}
            & 235 (+108\%) 
            & \cellcolor{best}28.46 & \cellcolor{best}0.834 & \cellcolor{best}0.182
            & \cellcolor{best}23.38 & 0.820 & 0.164
            & \cellcolor{best}29.56 & \cellcolor{best}0.898 & 0.193 \\
            Ours
            & 128 (+14\%)
            & \cellcolor{second}28.44 & \cellcolor{second}0.833 & \cellcolor{best}0.182
            & 23.28 & \cellcolor{second}0.822 & \cellcolor{best}0.158
            & \cellcolor{second}29.51 & 0.896 & \cellcolor{second}0.192 \\
        \bottomrule
    \end{tabular}
\end{table*}

%% file: figure/main.tex
\begin{figure*}[htbp]
    \centering
    \includegraphics[width=0.95\linewidth]{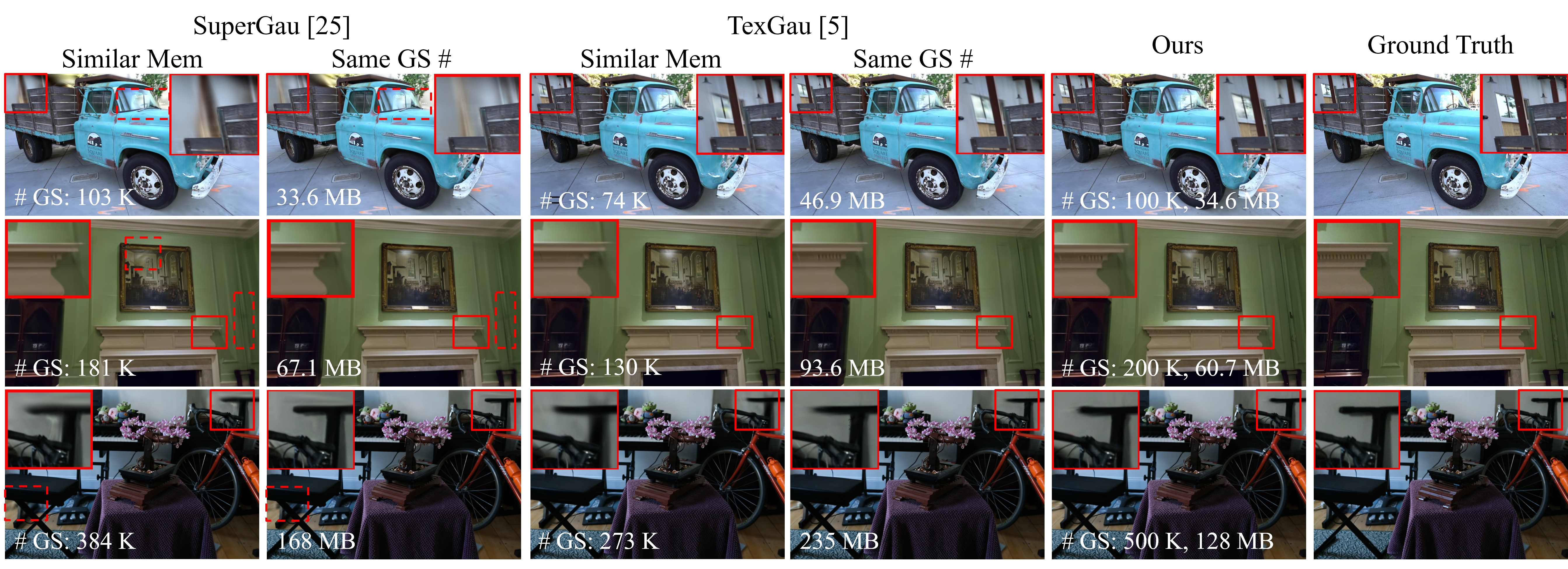} \\
    \vspace{-0.5em}
    \caption{\textbf{Qualitative Results.} Our method preserves rendering quality with far fewer parameters on a varied number of Gaussians across three datasets. With a similar model size, our methods achieve better rendering quality than baseline methods.}
    \label{fig:main_qualitative}
\end{figure*}

%% file: tab/ablate_warp.tex
\begin{table*}[t] \footnotesize
    \caption{\textbf{Ablation studies of the warping operations.} Experiments conducted with the equal number of Gaussians on the MipNeRF-360 dataset. Each pair of experiments uses the same initialization of 2D Gaussians and is trained for the same number of steps and time. The only differences within each pair lie in the adopted warping functions. The introduction of warping functions consistently improves the rendering quality under varied texture resolutions and number of Gaussians.}
    \vspace{-0.5em}
    \label{tab:ablation_warp}
    \centering
    \begin{tabular}{l lll lll lll}
        \toprule
        & \multicolumn{3}{c}{ \# 500K} &  \multicolumn{3}{c}{ \# 200K}&  \multicolumn{3}{c}{ \# 100K} \\
        \cmidrule(lr){2-4}  \cmidrule(lr){5-7}  \cmidrule{8-10}
        Method
        & PSNR $\uparrow$    & SSIM $\uparrow$   &  LPIPS $\downarrow$
        & PSNR $\uparrow$    & SSIM $\uparrow$   &  LPIPS $\downarrow$
        & PSNR $\uparrow$    & SSIM $\uparrow$   &  LPIPS $\downarrow$ \\
        \midrule
        TexGau\textsuperscript{\textdagger} (4x4)
        & 28.46 & 0.834 & 0.182
        & 27.81 & 0.813 & 0.223
        & 27.09 & 0.787 & 0.267 \\
        + axis warp
        & 28.56 \textcolor{red}{(+)} & 0.839 \textcolor{red}{(+)} & 0.171 \textcolor{red}{(-)}
        & 27.91 \textcolor{red}{(+)} & 0.819 \textcolor{red}{(+)} & 0.206 \textcolor{red}{(-)}
        & 27.21 \textcolor{red}{(+)} & 0.795 \textcolor{red}{(+)} & 0.244 \textcolor{red}{(-)} \\
        \midrule
        TexGau\textsuperscript{\textdagger} (6x6) 
        & 28.58 & 0.838 & 0.173
        & 27.98 & 0.819 & 0.209
        & 27.30 & 0.795 & 0.248 \\
        + axis warp
        & 28.65 \textcolor{red}{(+)} & 0.841 \textcolor{red}{(+)} & 0.162 \textcolor{red}{(-)}
        & 28.07 \textcolor{red}{(+)} & 0.824 \textcolor{red}{(+)} & 0.191 \textcolor{red}{(-)}
        & 27.45 \textcolor{red}{(+)} & 0.804 \textcolor{red}{(+)} & 0.223 \textcolor{red}{(-)} \\
        \bottomrule
    \end{tabular}
\end{table*}

%% file: tab/ablate_radial.tex
\begin{table*}[t]
\centering
\begin{minipage}{0.48\linewidth} \scriptsize
    \centering
    \caption{\textbf{Comparisons between axis-wise and radial CDF warping functions.} Experiments conducted with an equal number of Gaussians (500K) on Mip-NeRF 360.}\label{tab:ablation_radial_warp}
    \begin{tabular}{lccc}
        \toprule
        & PSNR $\uparrow$    & SSIM $\uparrow$   &  LPIPS $\downarrow$ \\
        \midrule
        TexGau\textsuperscript{\textdagger} (4x4) & 28.464 & 0.8345 & 0.1818 \\
        w/ axis warp & \cellcolor{second}28.558 & \cellcolor{second}0.8388 & \cellcolor{best}0.1708 \\
        w/ radial warp & \cellcolor{best}28.563 & \cellcolor{best}0.8390 & \cellcolor{second}0.1729 \\
        \midrule
        TexGau\textsuperscript{\textdagger} (6x6) & 28.580 & 0.838 & 0.173 \\
        w/ axis warp & \cellcolor{best}28.650 & \cellcolor{best}0.8414 & \cellcolor{best}0.1616 \\
        w/ radial warp & \cellcolor{second}28.633 &\cellcolor{second}0.8413 & \cellcolor{second}0.1641 \\
        \bottomrule
    \end{tabular}
\end{minipage}
\hfill
\begin{minipage}{0.48\linewidth} \small
    \centering
    \caption{\textbf{Performance of warping operations on RGB textures.} Experiments on Mip-NeRF 360 with varying Gaussian counts show that our warping operations consistently improve rendering quality, regardless of the underlying texture representation.}\label{tab:ablation_rgb}
    \begin{tabular}{llll}
        \toprule
        & PSNR $\uparrow$    & SSIM $\uparrow$   &  LPIPS $\downarrow$ \\
        \midrule
        TexGau\textsuperscript{\textdagger} (\# 200K) & 27.586 & 0.8071 & 0.2258 \\
        + axis warp & 27.642 \textcolor{red}{(+)} & 0.8093 \textcolor{red}{(+)} & 0.2189 \textcolor{red}{(-)} \\
        \midrule
        TexGau\textsuperscript{\textdagger} (\# 100K) & 26.845 & 0.7799 & 0.2712 \\
        + axis warp & 26.923 \textcolor{red}{(+)} & 0.7840 \textcolor{red}{(+)} & 0.2594 \textcolor{red}{(-)} \\
        \bottomrule
    \end{tabular}
\end{minipage}
\end{table*}

%% file: tab/effeciency.tex
\begin{table}[t]
\centering
    \caption{\textbf{Texture training time on MipNeRF-360.} Our method remains compatible with fast GPU training and adds only minor overhead.}\label{tab:time_training}
    \begin{tabular}{lccc}
        \toprule
        & \# 100K    & \# 200K   & \# 500K \\
        \midrule
        TexGau\textsuperscript{\textdagger} & 40m & 43m & 47m \\
        Ours & 42m & 46m & 52m \\
        \bottomrule
    \end{tabular}
\end{table}

\begin{table}[t]
    \centering
    \caption{\textbf{Rendering FPS on MipNeRF-360.} Experiments conducted with identical RGB textures to isolate alpha-blending differences. Warping introduces negligible cost, preserving real-time performance.}\label{tab:time_rendering}
    \begin{tabular}{llll}
        \toprule
        & \# 100K    & \# 200K   & \# 500K \\
        \midrule
        TexGau\textsuperscript{\textdagger} & 125.7 & 105.9 & 76.8 \\
        + axis warp & 120.5 & 104.4 & 75.4 \\
        + radial warp & 121.7 & 102.7 & 74.1 \\
        \bottomrule
    \end{tabular}
\end{table}

%% file: sec/5_conclusion.tex
\section{Conclusion.}
In this work, we present ASAP-Textured Gaussians, a novel approach to enhance texture representation in 2D Gaussian Splatting through adaptive sampling and anisotropic parameterization. By introducing warping functions that align textures with the underlying 2D Gaussian distributions, we enable efficient usage of texture resources. Additionally, our error-driven texture growth strategy dynamically adjusts texture resolutions based on learning status, allowing for detailed capture where necessary while avoiding over-allocation. Extensive experiments on multiple datasets demonstrate that our method achieves superior rendering quality with reduced model size compared to existing approaches.

%% file: sec/X_suppl.tex
\clearpage
\setcounter{page}{1}
\maketitlesupplementary


\counterwithin{figure}{section}
\counterwithin{table}{section}
\counterwithin{equation}{section}
\renewcommand\thesection{\Alph{section}}
\renewcommand\thetable{\thesection.\arabic{table}}
\renewcommand\thefigure{\thesection.\arabic{figure}}
\renewcommand\theequation{\thesection.\arabic{equation}}
\setcounter{section}{0}

\renewcommand{\theHsection}{supp.\Alph{section}}
\renewcommand{\theHtable}{supp.\Alph{section}.\arabic{table}}
\renewcommand{\theHfigure}{supp.\Alph{section}.\arabic{figure}}
\renewcommand{\theHequation}{supp.\Alph{section}.\arabic{equation}}

The supplementary material is organized as follows:
\begin{itemize}
\item \cref{sec:supp_derivations} presents detailed derivations of the two warping variants.
\item \cref{sec:supp_memory} reports additional experiments under fixed memory budgets.
\item \cref{sec:supp_results} includes further qualitative results.
\end{itemize}

\section{Warping function derivations}\label{sec:supp_derivations}
The goal of warping is to map canonical texture coordinates $\vu=(u,v)$ into a mass-aware texture domain $\tilde{\vu}=(\tilde{u},\tilde{v})$, such that a uniform sampling pattern in the warped domain corresponds to a Gaussian-weighted sampling pattern in the canonical domain. Formally, we desire the sampling density in canonical space to satisfy $P(\vu)\propto G(\vu)$, where texels should be denser near regions of higher Gaussian mass and sparser in low-mass regions.

For clarity, we first consider the 1D case with a scalar coordinate $u$ and warp $\tilde{u} = \phi(u)$.
The change of variables theorem gives
\[
P(u)\,\mathrm{d}u = \tilde{P}(\tilde{u})\,\mathrm{d}\tilde{u}
\quad\Rightarrow\quad
P(u) = \tilde{P}(\tilde{u})\,\left|\frac{\mathrm{d}\tilde{u}}{\mathrm{d}u}\right|.
\]
Given the uniform sampling in the warped domain, such that $\tilde{P}(\tilde{u}) = \mathrm{const}$, and our goal $P(u)\propto G(u)$, we obtain
\begin{equation}
\frac{\mathrm{d}\tilde{u}}{\mathrm{d}u} \propto G(u).
\end{equation}

Integrating this differential relationship produces the cumulative distribution function (CDF) based warping function:
\begin{equation}
\tilde{u} = \phi(u) = \int_{-\infty}^{u} G(t) \mathrm{d}t.
\end{equation}
Formally, this yields the closed form:
\begin{equation}
\tilde{u} = \phi(u) = \frac{1}{2} \left( 1 + \mathrm{erf}\left(\frac{u}{\sqrt{2}}\right) \right),
\end{equation}
which corresponds to our axis-wise warping in~\cref{eq:axis}.

\paragraph{Radial CDF Warping.}
One interesting property of the (normalized) canonical space of the 2D Gaussian is its radially symmetric structure.
Formally, the local density depends only on the radius $r = \lVert \vu \rVert$ from the center:
\begin{equation}
    \forall \, \lVert \vu_1 \rVert = r = \lVert \vu_2 \rVert, \quad G( \vu_1 ) = G( \vu_2 ) = G'(r),
\end{equation}
where $G'(r)\propto \exp(-r^{2}/2)$ denotes the radial profile of the isotropic standard Gaussian.

Similar to the 1D axis-wise case, we now work with the radial marginal density of $r=\lVert\vu\rVert$. In polar coordinates, the area element is $dx\,dy = r\,dr\,d\theta$, so the induced radial density is
\begin{equation}
    P(r) \propto r\,G'(r).
\end{equation}
With the constant density in the warped radial coordinate, $\tilde{P}(\tilde r)=\text{const}$, the 1D change-of-variables relation
\begin{equation}
    P(r)\,\mathrm{d}r \propto \tilde{P}(\tilde r)\,\mathrm{d}\tilde r
\end{equation}
gives
\begin{equation}
    r\,G'(r)\,\mathrm{d}r \propto \mathrm{d}\tilde r
    \quad\Rightarrow\quad
    \frac{\mathrm{d}\tilde r}{\mathrm{d}r} \propto r\,G'(r),
\end{equation}
whose integral yields the Rayleigh CDF
\begin{equation}
    \tilde r = \phi(r) = \int_{0}^{r} \tau\,G'(\tau)\,\mathrm{d}\tau
              = 1 - \exp\!\left(-\frac{r^{2}}{2}\right).
\end{equation}
This constitutes our radial CDF warping in~\cref{eq:radial}.

\section{Additional experiments under fixed memory budgets.}\label{sec:supp_memory}
\input{tab/supp_same_mem}
In this experiment, we control the number of Gaussians to ensure comparable overall memory consumption across methods. Although the Gaussian count affects both geometry fidelity and texture memory—and thus does not isolate texture efficiency—our method still achieves strong rendering quality under this controlled setting, as illustrated in~\cref{tab:supp_same_mem}.

\section{Qualitative results.}\label{sec:supp_results}
We include more qualitative results in~\cref{fig:supp_main_qualitative}.
\input{figure/supp_main}

%% file: tab/supp_same_mem.tex
\begin{table*}[t] \footnotesize 
    \caption{\textbf{Quantitative comparisons under fixed memory budgets.} 
    We control the number of Gaussians to match the overall memory usage across methods. Under this fixed memory budget, our approach in general delivers higher rendering quality than texture-based Gaussian baselines, highlighting the effectiveness of our design.
    }
    \label{tab:supp_same_mem}
    \centering
    \begin{tabular}{l l ccc ccc ccc}
        \toprule
        Mem &  & \multicolumn{3}{c}{ Mip-NeRF 360} &  \multicolumn{3}{c}{ Tanks \& Temples}&  \multicolumn{3}{c}{ DeepBlending} \\
        \cmidrule(rr){1-1} \cmidrule(lr){2-2} \cmidrule(lr){3-5}  \cmidrule(lr){6-8}  \cmidrule{9-11}
        Method
        & \scriptsize \# Gaussians
        & \scriptsize PSNR $\uparrow$    & \scriptsize SSIM $\uparrow$  & \scriptsize LPIPS $\downarrow$
        & \scriptsize PSNR $\uparrow$    & \scriptsize SSIM $\uparrow$  & \scriptsize LPIPS $\downarrow$
        & \scriptsize PSNR $\uparrow$    & \scriptsize SSIM $\uparrow$  & \scriptsize LPIPS $\downarrow$
        \\
        \midrule
        \multicolumn{2}{l}{$\approx$ 34.6MB} \\
        \midrule
            2DGS\textsuperscript{\textdagger}
            & 154K
            & \cellcolor{second}27.05 & \cellcolor{best}0.791 & \cellcolor{best}0.257
            & \cellcolor{second}22.48 & \cellcolor{best}0.793 & \cellcolor{second}0.217
            & \cellcolor{second}28.95 & \cellcolor{second}0.889 & \cellcolor{second}0.230 \\
            SuperGau
            & 103K
            & 24.04 & 0.731 & 0.351
            & 20.59 & 0.740 & 0.291
            & 26.74 & 0.873 & 0.282 \\
            TexGau\textsuperscript{\textdagger}
            & 74K 
            & 26.64 & 0.772 & 0.292
            & 22.34 & 0.776 & 0.251
            & 28.93 & 0.886 & 0.249 \\
            Ours
            & 100K
            & \cellcolor{best}27.11 & \cellcolor{second}0.785 & \cellcolor{second}0.259
            & \cellcolor{best}22.64 & \cellcolor{best}0.793 & \cellcolor{best}0.216
            & \cellcolor{best}29.12 & \cellcolor{best}0.890 & \cellcolor{best}0.227 \\
        \arrayrulecolor{black}
        \midrule
        \multicolumn{2}{l}{$\approx$ 60.7MB} \\
        \midrule
            2DGS\textsuperscript{\textdagger}
            & 270K 
            & \cellcolor{second}27.75 & \cellcolor{best}0.811 & \cellcolor{second}0.223 
            & \cellcolor{second}22.85 & \cellcolor{second}0.806 & \cellcolor{second}0.191
            & 29.25 & \cellcolor{second}0.892 & \cellcolor{second}0.214 \\
            SuperGau 
            & 181K 
            & 25.44 & 0.772 & 0.283 
            & 21.87 & 0.788 & 0.221
            & 28.65 & 0.888 & 0.232 \\
            TexGau\textsuperscript{\textdagger} 
            & 130K
            & 27.31 & 0.797 & 0.250
            & 22.76 & 0.796 & 0.218
            & \cellcolor{second}29.29 & \cellcolor{second}0.892 & 0.229 \\
            Ours
            & 200K
            & \cellcolor{best}27.79 & \cellcolor{second}0.810 & \cellcolor{best}0.220
            & \cellcolor{best}22.97 & \cellcolor{best}0.809 & \cellcolor{best}0.186
            & \cellcolor{best}29.40 & \cellcolor{best}0.893 & \cellcolor{best}0.208 \\
        \midrule
        \multicolumn{2}{l}{$\approx$ 128MB} \\
        \midrule
            2DGS\textsuperscript{\textdagger}
            & 570K 
            & \cellcolor{second}28.26 & \cellcolor{second}0.832 & \cellcolor{second}0.185
            & 23.15 & 0.819 & \cellcolor{second}0.164
            & 29.21 & 0.894 & 0.198 \\
            SuperGau 
            & 384K 
            & 27.17 & 0.806 & 0.212
            & \cellcolor{second}23.21 & \cellcolor{best}0.824 & 0.166 
            & 29.41 & \cellcolor{second}0.896 & \cellcolor{second}0.195 \\
            TexGau\textsuperscript{\textdagger} 
            & 273K 
            & 28.09 & 0.822 & 0.207
            & 23.17 & 0.814 & 0.183
            & 29.47 & \cellcolor{best}0.897 & 0.204 \\
            Ours
            & 500K
            & \cellcolor{best}28.44 & \cellcolor{best}0.833 & \cellcolor{best}0.182
            & \cellcolor{best}23.28 & \cellcolor{second}0.822 & \cellcolor{best}0.158
            & \cellcolor{best}29.51 & \cellcolor{second}0.896 & \cellcolor{best}0.192 \\
        \bottomrule
    \end{tabular}
\end{table*}

%% file: figure/supp_main.tex
\begin{figure*}[htbp]
    \centering
    \includegraphics[width=0.95\linewidth]{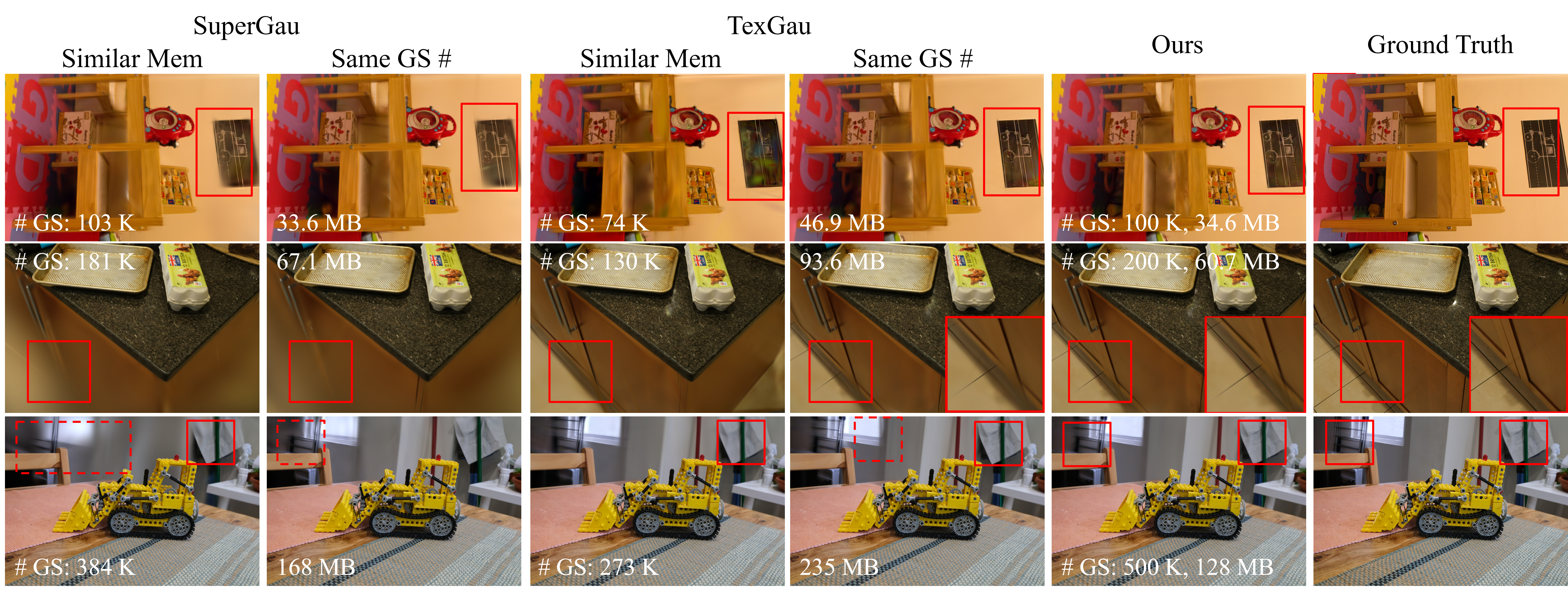} \\
    \vspace{-0.5em}
    \caption{\textbf{More Qualitative Results.} Our method preserves rendering quality with far fewer parameters on a varied number of Gaussians across three datasets. With a similar model size, our methods achieve better rendering quality than baseline methods.}
    \label{fig:supp_main_qualitative}
\end{figure*}